%% file: main.tex
\pdfoutput=1

\documentclass[11pt]{article}

\usepackage[final]{acl}

\usepackage{times}
\usepackage{latexsym}
\usepackage{balance}

\usepackage[T1]{fontenc}

\usepackage[utf8]{inputenc}

\usepackage{microtype}

\usepackage{inconsolata}

\usepackage{graphicx}

\usepackage{microtype}
\usepackage{hyperref}
\usepackage{url}
\usepackage{booktabs}
\usepackage{bm}
\usepackage{xspace}
\definecolor{darkblue}{rgb}{0, 0, 0.5}
\hypersetup{colorlinks=true, citecolor=darkblue, linkcolor=darkblue, urlcolor=darkblue}
\usepackage{adjustbox}

\usepackage{amsmath}
\usepackage{amssymb}
\usepackage{amsfonts}
\usepackage{multicol}
\usepackage{mathtools}

\usepackage{array}

\usepackage{here}

\def\eg{{\it e.g.}}

\def\ie{{\it i.e.}}

\title{Text2Traj2Text: Learning-by-Synthesis Framework for\\Contextual Captioning of Human Movement Trajectories}

\author{ \\
    \textbf{Hikaru Asano${}^{1,2}$\thanks{Work done during an internship at CyberAgent Inc.} $\;\;\;$ Ryo Yonetani${}^3$ $\;\;\;$ Taiki Sekii${}^3$ $\;\;\;$ Hiroki Ouchi${}^{4,3,2}$} \\
    ${}^1$The University of Tokyo \quad
    ${}^2$RIKEN AIP \quad
    ${}^3$CyberAgent Inc. \quad \\
    ${}^4$Nara Institute of Science and Technology
    \\
    \texttt{asano-hikaru19@g.ecc.u-tokyo.ac.jp,} \\
    \texttt{\{yonetani\_ryo, sekii\_taiki\}@cyberagent.co.jp,} \\
    \texttt{hiroki.ouchi@is.naist.jp} \\
}

\usepackage{graphicx}
\usepackage{listings}
\usepackage{tcolorbox}
\usepackage{multirow}

\newcommand{\methodname}{\textsc{Text2Traj2Text}\xspace}

\begin{document}

\maketitle

\begin{abstract}
This paper presents Text2Traj2Text, a novel learning-by-synthesis framework for captioning possible contexts behind shopper's trajectory data in retail stores. Our work will impact various retail applications that need better customer understanding, such as targeted advertising and inventory management. The key idea is leveraging large language models to synthesize a diverse and realistic collection of contextual captions as well as the corresponding movement trajectories on a store map. Despite learned from fully synthesized data, the captioning model can generalize well to trajectories/captions created by real human subjects. Our systematic evaluation confirmed the effectiveness of the proposed framework over competitive approaches in terms of ROUGE and BERT Score metrics.
\end{abstract}

\input{contents/1_intro}

\input{contents/2_prelim}

\input{contents/3_method}

\input{contents/4_experiment_rearranged}
\balance
\input{contents/6_related_works.tex}

\input{contents/7_conclusion}

\section*{Acknowledgments}
We would like to express our gratitude for the anonymous reviewers who provided many insightful comments that have improved our paper.
Special thanks also go to Akira Kasuga and the other members of CyberAgent, Inc. AI Lab for the interesting comments, energetic discussions, and valuable code review.

\clearpage
\bibliography{library}

\clearpage
\appendix
\input{contents/appendix}

\end{document}

%% file: contents/1_intro.tex
\section{Introduction}
Retail is an essential industry that is closely tied to our daily lives. Imagine a customer visiting a supermarket. The customer first goes to the fruit section and compares various products. Next, they go to the fish section, where they compare two products. Afterward, they pass by the processed food section and head to the checkout, purchasing discounted organic strawberries and fish. From these movements and purchases, one can guess that ``\emph{the customer is budget-conscious and interested in healthy meals.}''

\begin{figure}[t]
    \centering
    \includegraphics[width=\linewidth]{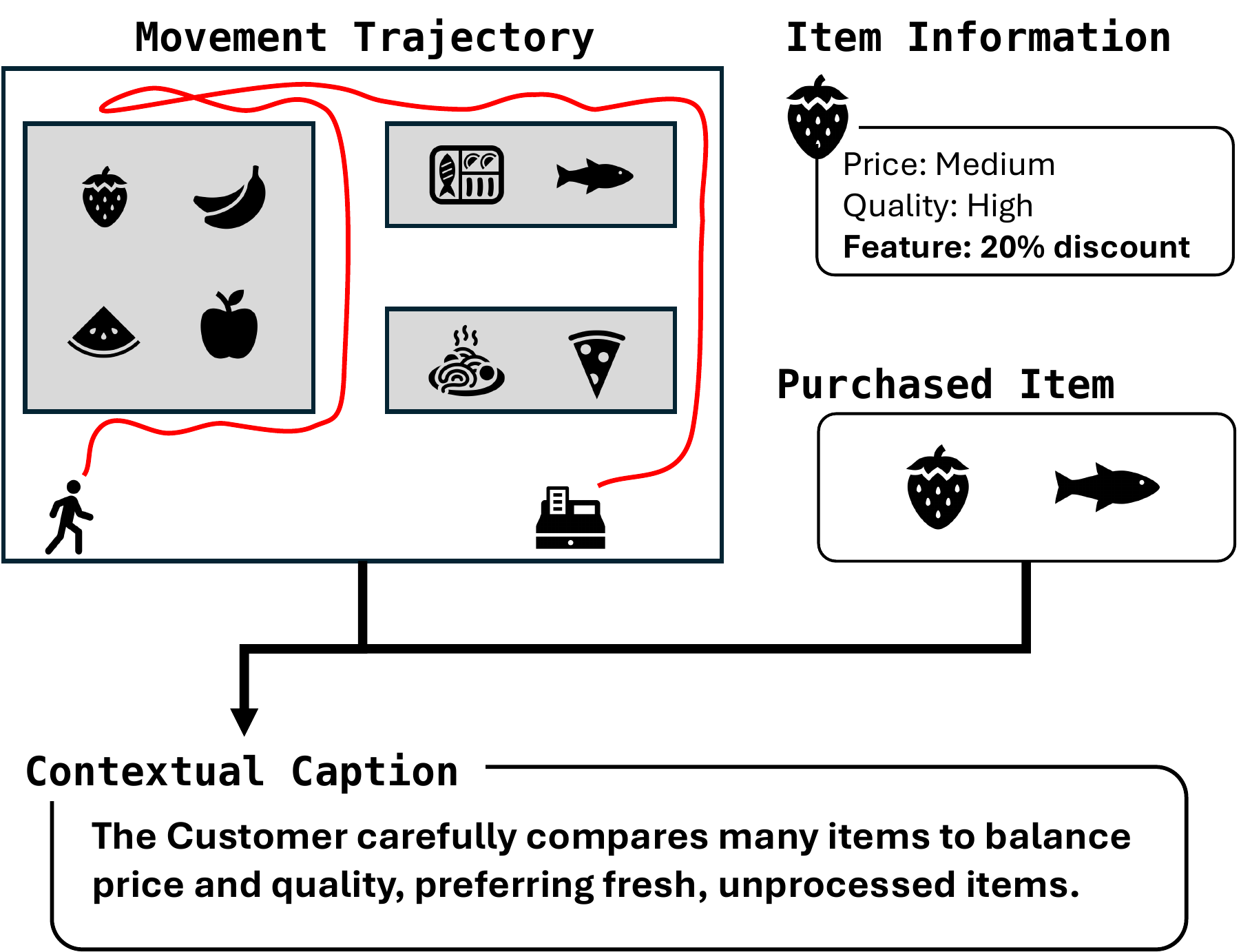}
    \caption{\textbf{Contextual Captioning of Human Movement Trajectories.} Given a human movement trajectory associated with semantic information such as nearby items and actual purchases in a retail store, we aim to produce contextual captions that best explain the possible contexts behind.}
    \label{fig:teaser}
  \end{figure}

Such profiling and verbalization of possible contexts behind shopping behaviors is vital for retailers to improve customer understanding and customer experience. We are interested in automating this intelligent activity, with recent advances in large-scale language modeling. Doing so would help facilitate and scale up retailer's operations beyond the number of experts, and can also enhance several applications such as targeted advertising~\cite{liu2018tar,ghose2019mobile} and inventory management~\cite{carreras2013store}.

As the first step toward this goal, we formulate a new task, \emph{contextual captioning of human movement trajectories}, with a particular focus on retail applications. Let us illustrate an example in Fig.~\ref{fig:teaser}. The input of this task is a movement trajectory associated with its semantic information, such as item positions and purchased items for a customer navigating a retail store. The output is a \emph{contextual caption} that explains a possible context behind the demonstrated trajectories, such as purposes and preferences for the purchases.

While it is intuitive to learn neural captioning models for this task, it is nontrivial how to gather the sufficient number of training data, more specifically trajectories annotated with contextual captions. Although recent advancements in wireless sensing technologies have already enabled accurate indoor localization~\cite{zafari2019survey}, collecting actual customer locations in stores is often nontrivial due to privacy concerns. Even if location data were available, annotating appropriate captions for them is labor intensive.

In this work, we present \textsc{Text2Traj2Text}, a learning-by-synthesis framework to address this challenge. As illustrated in Fig.~\ref{fig:overview}, this framework consists of two phases: \textsc{Text2Traj} (data synthesis) and \textsc{Traj2Text} (model fine-tuning). In the \textsc{Text2Traj} phase, we leverage large language models (LLMs) to synthesize realistic and diverse collections of contextual captions as well as concrete trajectories on store maps. Then in the \textsc{Traj2Text} phase, we construct a captioning model fine-tuned on the synthesized data.

Through systematic evaluation, we show that the diverse data synthesis by LLMs allows our captioning model to generalize well to actual human trajectories and human-created captions. It also outperforms several existing LLM services (GPT-3.5~\citep{openai2023chatgpt}, GPT-4~\citep{openai2023gpt4}) as well as open-source benchmark Llama2~\citep{llama2} adapted to the task via in-context learning, in terms of ROUGE and BERT Score metrics.

Our contributions are summarized as follows: (1)~formulating a new captioning task called contextual captioning of human movement trajectories; (2)~proposing a learning-by-synthesis framework, \textsc{Text2Traj2Text}, and demonstrating its effectiveness on actual human data; (3)~creating a benchmark dataset to accelerate future research.\footnote{Our code and dataset will be available at \url{https://github.com/CyberAgentAILab/text2traj2text}.}

%% file: contents/2_prelim.tex
\section{Contextual Captioning of Human Movement Trajectories}
\label{sec:task}

\subsection{Motivating Scenario}
The goal of our task is to generate concise text that describes possible underlying contexts of human movement trajectories, such as purposes and preferences. We focus particularly on a retail scenario, where people walk around a store, browse items of interest, and choose some to buy. Retailers analyze such shopping behaviors collected from consenting customers to gain deeper understanding of customers and improve customer experiences via demand prediction, inventory management, or targeted advertising. Much like web search engines automatically infer user preferences from click streams, we aim to automate customer activity profiling, ultimately across a wide range of stores beyond what is possible with a limited number of experts. Formatting profile results as sentences, as human experts do when communicating with stakeholders, is crucial for improving the interpretability of such automation.

\subsection{Task Formulation}
Given a movement trajectory $X$ and its semantics including \emph{items in contact} $I$ and \emph{purchased items} $\mathcal{P}$, we aim to generate a contextual caption $S$, as each detailed below.
\paragraph{Input: Trajectory and its semantics.}
The movement trajectory is a sequence of $T$ locations, \ie, $X = (x_1, \dots, x_T)$, where $x_t\in\mathbb{R}^2$ corresponds to a 2-D location where the customer stayed at each time step $t$. \emph{Items in contact} are the items closest to the customer at each time step, \ie, ${I}=({i}_1,\dots {i}_T)$. \emph{Purchased items} are the items that the customer purchased, which form a subset of the items in contact, \ie,  $\mathcal{P}\subset I$. Technically, it is feasible to collect those data from consenting customers via wireless indoor localization technologies~\cite{zafari2019survey} used in combination with point-of-sales (POS) systems. Nevertheless, such data collection is hard to scale in practice, as it is difficult to intervene in a retail store currently operating and obtain approval from each customer.
\paragraph{Output: Contextual captions.}
The \emph{contextual caption} is a sequence of tokens, \ie, $S = (s_1, s_2, \dots )$, where $s$ is a token. We assume that each caption is concise, typically spanning a few sentences, and describes various aspects of the customer's shopping behavior such as their preferences for price versus quality, required quantity, and other characteristics related to item choices (\eg, ready-to-eat, health-conscious).

%% file: contents/3_method.tex
\section{\methodname}

\begin{figure*}[t]
    \centering
    \includegraphics[width=\textwidth]{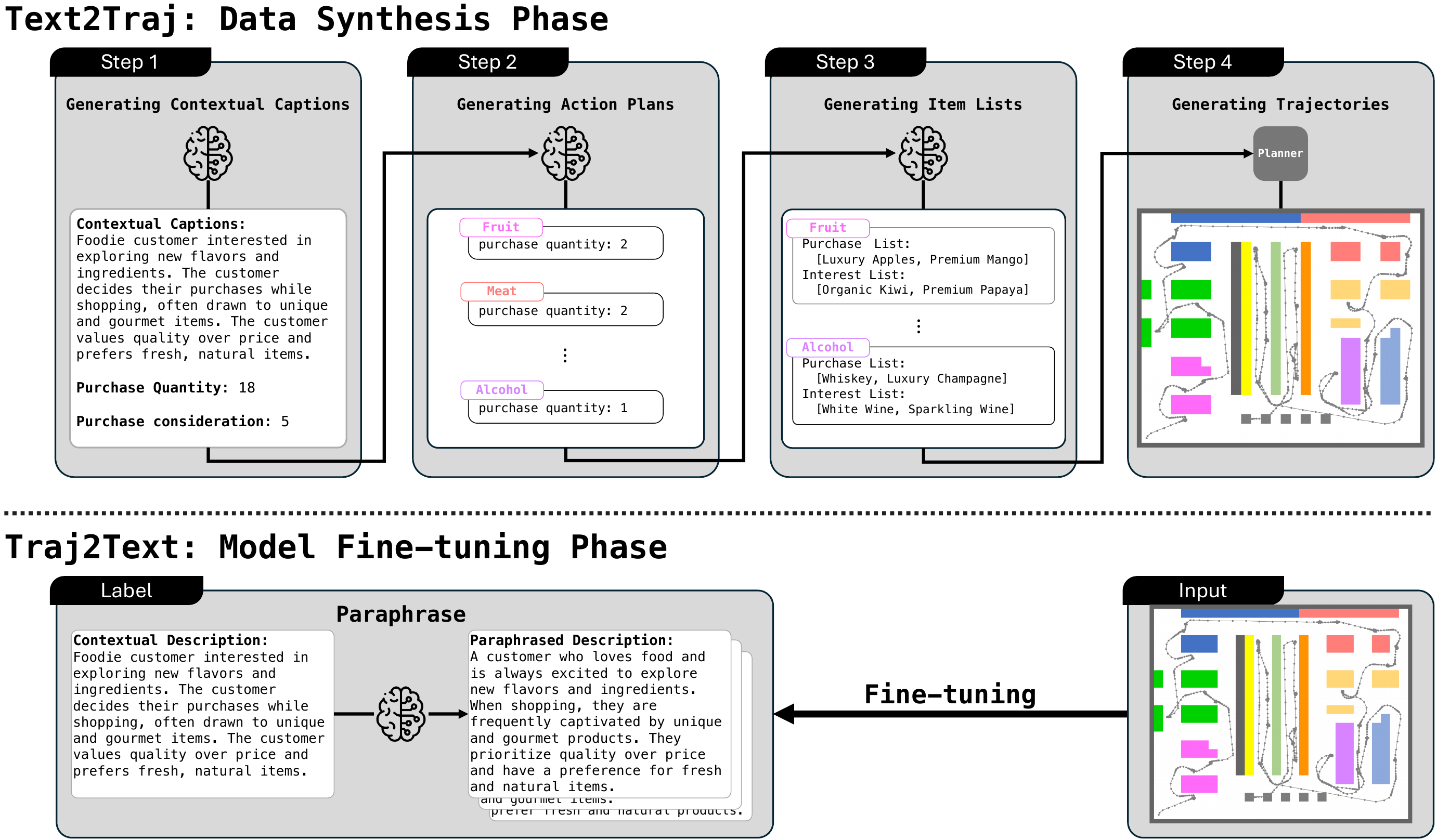}
    \caption{\textbf{Text2Traj2Text Framework}. (1) \textsc{Text2Traj}: We leverage LLMs to synthesize contextual captions and their instances as concrete action plans, item lists, and in-store trajectories. (2) \textsc{Traj2Text}: We fine-tune a language model with the synthesized data to be able to produce contextual captions from trajectory data.}
    \label{fig:overview}
\end{figure*}

Fig.~\ref{fig:overview} illustrates the overview of the proposed framework, \textsc{Text2Traj2Text}, which consists of \textsc{Text2Traj} data synthesis phase and \textsc{Traj2Text} model fine-tuning phase.

\subsection{\textsc{Text2Traj}: Data Synthesis}
\label{sec:text2traj}
In the \textsc{Text2Traj} phase, we propose leveraging pretrained, instruction-tuned LLMs in combination with a human trajectory planner to synthesize a diverse and realistic collection of annotated trajectory data. This approach is inspired by recent advancements in robotics research that aim to generate complex robot motion by incorporating LLMs into hierarchical motion planning frameworks~\citep{ahn2022can,wang2023gensim,wang2023describe,liu2023llm+}. It utilizes the reasoning ability of LLMs for task planning to determine which actions to take or which goals to approach, while employing classical motion planning to generate feasible motion trajectories for each action. Similarly, in our framework, an LLM first creates diverse contextual captions and instantiates coarse action plans from the captions. A trajectory planner then traces the plans to generate feasible movement trajectories on a store map. More specifically, the \textsc{Text2Traj} phase consists of four steps as shown below.

\paragraph{Step 1: Generating contextual captions.}
First, we give a prompt (Fig.~\ref{fig:prompt-step1} in Appendix~\ref{sec:prompt}) to an LLM for producing contextual captions on three types of information: (i) individual customer's product preferences (\eg, \emph{``loves apple''}), (ii) category-level interests (\eg, \emph{``interested in fruits''}), and (iii) decision-making tendencies (\eg, \emph{``have a list of items to purchase''}). The LLM's output also includes the number of items planned to purchase (\ie, purchase quantity) and the person's purchase consideration. Higher purchase consideration suggests more comparison of products before purchasing, while a lower one indicates a tendency to have pre-determined shopping plan. 

\paragraph{Step 2: Generating action plans.}
Given a prompt (Fig.~\ref{fig:prompt-step2} in Appendix~\ref{sec:prompt}) that contains the outputs from Step 1 (\ie, a contextual caption and purchase quantity) and item categories in a store, the LLM generates an action plan, a list of pairs of item categories and their expected purchase quantity, \eg, \emph{\{``fruit'': 4, ``meat'': 0, ``alcohol'': 1\}}.

\paragraph{Step 3: Generating item lists.}
Given a prompt (Fig.~\ref{fig:prompt-step3} in Appendix~\ref{sec:prompt}), the LLM converts each item category determined in Step 2 into more specific item information, \ie, (i) a \emph{purchase list} consisting of the name of items planned to purchase, and (ii) an \emph{interest list} of items that the individual is likely to show interest in. The interest item will contain more items as the purchase consideration is set higher. Also, the number of items in the purchase list may not always match the planned purchase quantity generated in the previous step, as the number of actual purchases can change based on other factors, such as the availability of suitable items in the store.

\paragraph{Step 4: Generating movement trajectories.}
Finally, based on the purchase and interest lists generated in Step 3, we invoke a trajectory planner to instantiate concrete human movement trajectories on a store map.
We first assign ranks to each item category stochastically for each trajectory generation, with the rank reflecting the category's relative position in the store layout. The rank tendency is predefined based on the store's layout, where categories located closer to the entrance typically receive a higher rank.

The purchase consideration is again considered here; if it is set high, ranks have higher variances, resulting in more exploratory behaviors. Starting from a fixed starting location $x_0 \in \Omega$ (\eg, the entrance), the planner generates a feasible trajectory traversing items in the purchase and interest lists according to the category ranks in a store map like the one shown in Fig.~\ref{fig:teaser}.

\subsection{\textsc{Traj2Text}: Model Fine-tuning}
\label{sec:traj2text}
In the \textsc{Text2Traj} phase introduced so far, we first synthesize contextual captions and then instantiate concrete trajectories.
Reversely, in the following \textsc{Traj2Text} phase, we aim to build a captioning model that takes the synthesized trajectory data as input to produce plausible captions.

\paragraph{Input translation.}
As the input to the captioning model, we translate movement trajectory $X=(x_1, \dots, x_T)$, items in contact $I=(i_1,\dots,i_T)$, and purchased items $\mathcal{P}$, into textual representations. Importantly, movement trajectories can become lengthy as customers take more time for shopping, and can also contain many mundane moments. Here, we adopt a simple yet effective filtering technique to focus on important events in the trajectories. First, we calculate the displacement between consecutive locations, \ie, $\|x_t - x_{t-1}\|$, and extract moments when the individual stopped based on if the displacements are below a predetermined threshold. Then, items in contact at the stopping moments as well as those in the purchase list are simply concatenated: ``\texttt{Trajectory is fruit</s>vegetable</s> ...\textbackslash n Customer purchase item list is  ['Carrots', 'Beef'...] \textbackslash n Output:}.''

\paragraph{Data augmentation.}
The diversity of training data is crucial for the high generalization capability of learned models. While synthesized trajectories can sufficiently be diversified based on randomized ranks of item categories (in Step 4 of Sec.~\ref{sec:text2traj}), the variety of contextual captions may still be limited due to the expressiveness of the used LLM. To ensure high diversity for the captions, we introduce \emph{data augmentation by paraphrasing}; for each annotated trajectory, we let the LLM to produce alternative expressions of the caption with similar meanings, and relabel the trajectory accordingly.

%% file: contents/4_experiment_rearranged.tex
\section{Experiments}
We conducted systematic experiments to evaluate the effectiveness of the \methodname framework. Through the experiments, we aim to answer the following questions:
\begin{description}
    \item[\lbrack RQ.1\rbrack] Can the models trained by our proposed framework generate appropriate captions for synthesized trajectories? (Sec.~\ref{sec:exp-syn})
    \item[\lbrack RQ.2\rbrack] Can the models generalize to human-created trajectories/captions? (Sec.~\ref{sec:exp-human})
    \item[\lbrack RQ.3\rbrack] Can the models generalize to unseen maps? (Sec.~\ref{sec:exp-human})
\end{description}

\subsection{Experimental Setup}
\label{sec:exp_setup}
\paragraph{Data synthesis.}
Following Sec.~\ref{sec:text2traj}, we synthesized 80 pairs of contextual captions and the corresponding movement trajectories using GPT-4~\citep{openai2023gpt4}, while assuming a scenario of shopping at a supermarket. See Appendix~\ref{sec:prompt} for concrete prompts and Tab.~\ref{tab:qualitative_result} for the store map we used. We adopted a classical hierarchical planning framework for trajectory generation; a global planner (probabilistic roadmaps proposed by \citet{kavraki1996probabilistic}) first determines a sequence of sub-goals from the current item to the next one, and a local planner (dynamic window approach proposed by \citet{fox1997dynamic}) then produces a collision-free trajectory between the sub-goals. The synthesized data were divided into 64 training and 16 validation samples and augmented by paraphrasing with GPT-3.5~\cite{openai2023chatgpt}, where the number of added captions from a single original caption was 2, 4 or 8. For example, in the case of adding 8 paraphrases, the total number of training samples becomes $64 \times 9$ (where $1$ is the original caption and $8$ is its paraphrased captions).\footnote{Synthesizing captions is more complex than paraphrasing them, where we adopted GPT-4 for the former task and GPT-3.5 for the latter to consider cost-effectiveness.}

\paragraph{Implementation details.}
On the synthesized data, we fine-tuned the T5-Base model~\citep{10.5555/3455716.3455856} available on HuggingFace\footnote{\url{https://huggingface.co/t5-base}}, as its encoder-decoder structure was demonstrated effective for multimodal generation tasks~\citep{Xu2023-vz}. All fine-tuning was conducted on a single Tesla T4 GPU using AdamW optimizer with a learning rate of $5.6\times 10^{-5}$, where the batch size and the number of epochs were set to 8 and 5, respectively. The model checkpoint with the BERT Precision score~\citep{zhang2020bertscore} highest for the validation data was used for evaluation.

\paragraph{Baseline models.}
We compared our captioning model against the following baselines:
(a) T5-Small and T5-Base~\citep{10.5555/3455716.3455856} fine-tuned without paraphrasing-based data augmentation; (2) GPT-3.5~\citep{openai2023chatgpt}, GPT-4~\citep{openai2023gpt4}, and the open-source benchmark Llama-2-7b-chat-hf (referred to as Llama2)~\citep{llama2}\footnote{\url{https://huggingface.co/meta-llama/Llama-2-7b-chat-hf}}. GPT-3.5, GPT-4, and Llama2 were used via in-context learning; following \cite{maynezBenchmarkingLargeLanguage2023}, a few (1, 2, or 4) samples randomly selected from the training data were given as examples, and contextual captions were generated for the given movement trajectory.

\paragraph{Evaluation metrics.}
We employed ROUGE (R-1, R-2, R-L)~\citep{lin-2004-rouge} and BERT Score (BS-precision, recall, f1 score)~\citep{zhang2020bertscore} as evaluation metrics. ROUGE score captures lexical overlap by comparing n-grams and word sequences between generated and reference texts, while BERT Score, which utilizes BERT embeddings, measures semantic similarity.

\subsection{Evaluation with Synthesized Trajectories}
\label{sec:exp-syn}

\paragraph{[RQ.1] Can the models trained by our proposed framework generate appropriate captions for synthesized trajectories?}
Tab.~\ref{tab:sim_result} presents the quantitative results on 20 synthesized trajectories created in the same way as training/validation data. Overall, our model achieved the best performance even with an order-of-magnitude fewer parameters (223M) compared to the GPT family and Llama2 (over billions). We observe a monotonic improvement in nearly all metrics as the number of paraphrases increases, indicating the effectiveness of our data augmentation strategy. In contrast, T5-Small and T5-Base with vanilla fine-tuning demonstrated quite limited performances. The number of examples presented to GPT-3.5, GPT-4, and Llama2 was critical for their in-context learning ability, but this comes with increased inference costs and limits practical scalability.

\begin{table}[t]
    \centering
\scalebox{0.7}{
    \begin{tabular}{lcccccc}
        \toprule
        Models &  R-1 & R-2 & R-L & BS-p & BS-r & BS-f1\\
        \midrule
        T5-Small & 0.069 & 0.015 & 0.060 & 0.792 & 0.770 & 0.816 \\
        T5-Base& 0.287 & 0.094 & 0.243 & 0.860 & 0.859 & 0.861 \\
        \midrule
        GPT-3.5 & 0.240 & 0.049 & 0.151 & 0.854 & 0.841 & 0.868 \\
        \ \ + 1 examples & 0.326 & 0.080 & 0.211 & 0.887 & 0.883 & 0.891 \\
        \ \ + 2 examples & 0.358 & 0.093 & 0.225 & 0.892 & 0.888 & 0.895 \\
        \ \ + 4 examples & 0.364 & 0.101 & 0.235 & 0.894 & 0.890 & 0.897 \\
        \midrule
        GPT-4 & 0.180 & 0.034 & 0.119 & 0.844 & 0.822 & 0.868 \\
        \ \ + 1 examples & 0.322 & 0.064 & 0.192 & 0.881 & 0.873 & 0.890 \\
        \ \ + 2 examples & 0.334 & 0.070 & 0.199 & 0.887 & 0.881 & 0.894 \\
        \ \ + 4 examples & 0.378 & 0.106 & 0.240 & 0.897 & 0.892 & 0.902 \\
        \midrule
        Llama2 & 0.199 & 0.020 & 0.129 & 0.819 & 0.788 & 0.854 \\
        \ \ + 1 examples & 0.255 & 0.070 & 0.167 & 0.834 & 0.790 & 0.885 \\
        \ \ + 2 examples & 0.305 & 0.089 & 0.198 & 0.855 & 0.824 & 0.889 \\
        \ \ + 4 examples & 0.391 & 0.128 & 0.267 & 0.886 & 0.877 & 0.897 \\
        \midrule
        Ours &  &  &  &  &  &  \\
        \ \  2 paraphrases  & 0.374 & \textbf{0.140} & \textbf{0.297} & 0.888 & 0.894 & 0.882 \\
        \ \  4 paraphrases  & 0.368 & 0.131 & 0.287 & 0.888 & 0.893 & 0.884 \\
        \ \  8 paraphrases  & \textbf{0.412} & 0.138 & \textbf{0.297} & \textbf{0.907} & \textbf{0.910} & \textbf{0.905} \\
        \bottomrule
    \end{tabular}
    }
    \caption{Quantitative results for synthesized data.}
    \label{tab:sim_result}
\end{table}

\paragraph{Ablation study.}
Additionally, we investigate how each of the movement trajectories (with the list of nearby items) and the purchased items can contribute to the final performances using the validation dataset. In Tab.~\ref{tab:ablation}, we evaluated the following degraded variants: \textbf{w/o Traj} (resp. \textbf{w/o Item}) that removed trajectories (resp. purchased items) from the input; \textbf{w/ Shuffle Traj} (resp. \textbf{w/ Shuffle Item}) that replaced trajectories (resp. items) with those of other samples dataset according to the permutation feature importance method~\citep{breiman2001random, Fisher2019-hb}. These degraded versions all demonstrated quite limited performances, indicating the necessity of combining trajectories and purchased items for inferring contexts reliably. We also evaluate a more challenging case when the trajectory data are partially perturbed, possibly due to the inaccuracy of indoor localization systems. Our model is robust to such noises, as shown in the table (\textbf{w/ 5\% noise}).

\begin{table}[t]
    \centering
\scalebox{0.7}{
    \begin{tabular}{lllllll}
        \toprule
        Models &  R-1 & R-2 & R-L & BS-p & BS-r & BS-f1\\
        \midrule
        w/o Traj & 0.337 & 0.101 & 0.234 & 0.877 & 0.874 & 0.880 \\
        w/o Item & 0.218 & 0.038 & 0.166 & 0.862 & 0.876 & 0.849 \\
        w/ Shuffle Traj & 0.395 & 0.130 & 0.277 & 0.901 & 0.904 & 0.899 \\ 
        w/ Shuffle Item & 0.382 & 0.116 & 0.269 & 0.899 & 0.902 & 0.897 \\
        w/ 5\% noise & 0.428 & 0.159 & 0.308 & 0.907 & 0.911 & 0.903 \\
        \midrule
        Ours  &  0.427 & 0.156 & 0.308 & 0.907 & 0.911 & 0.903 \\
        \bottomrule
    \end{tabular}
    }
    \caption{Ablation study and noisy robustness evaluation}
    \label{tab:ablation}
\end{table}

\subsection{Evaluation with Real Human Data}
\label{sec:exp-human}

\paragraph{Data collection from human subjects.}
\begin{figure*}[th]
    \centering
    \includegraphics[width=\textwidth]{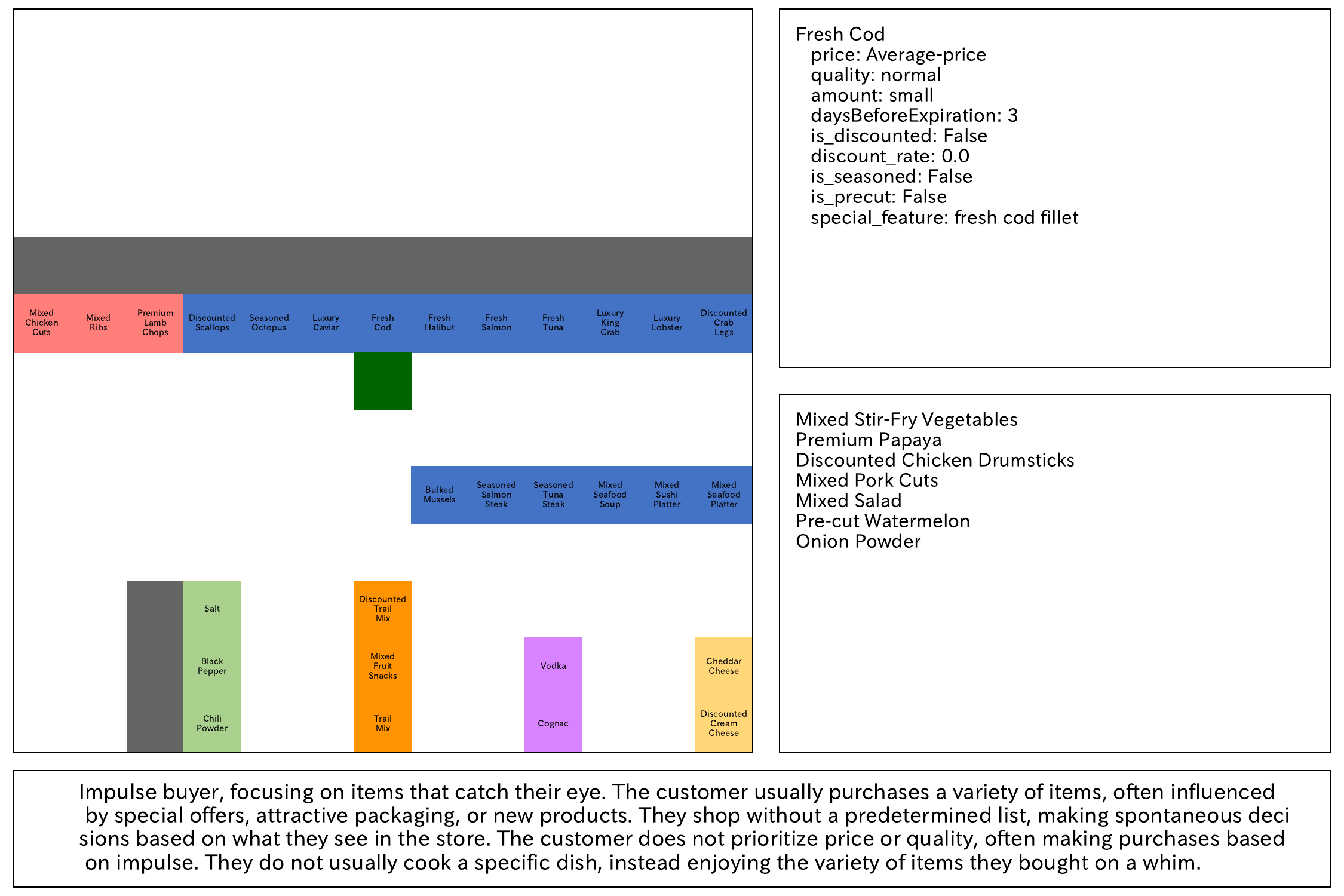}
    \caption{Visual user interface used to collect human-created trajectories. The green square represents the current position. Information on the closest item is shown in the upper right corner, and the list of items added to the cart is shown in the lower right corner. The caption to be followed is presented at the bottom of the screen.}
    \label{fig:game_environment}
\end{figure*}

We recruited eight participants to collect real human data for our study. The entire experiment consisted of two phases with different tasks.
In the first phase, two participants were instructed to create four plausible contextual captions about supermarket shoppers. Before they began, we provided them with three example captions to ensure appropriateness for our task.
In the second phase, six participants were asked to create trajectories using a visual interface (Fig.~\ref{fig:game_environment}) based on 10 randomly selected captions --- half synthetic and half created by the participants in the first phase. On the visual interface, the current position of a participant was marked by a green rectangle, with details about the item adjacent to their current location shown in the top right corner and items currently added to their cart displayed in the bottom right. Participants were allowed to navigate in the store map and add or remove adjacent items from their cart using keyboard input. Each session began from a fixed location and ended when participant reached the cashier register, tracking whole movement trajectories and final purchases.

Two distinct store maps were adopted in the experiment to validate the generalization ability of trained models: one used for training data and another as a completely new environment. Participants first completed two pilot rounds on one map to familiarize themselves with the interface and layout, followed by five main rounds on this map for data collection. They then repeated the same process on the other map. The set and order of captions, as well as store maps, were randomized across participants. Each experiment lasted about one hour. Through this experimental procedure, we collected 60 sufficiently diverse trajectory data points from real humans, as summarized in Tab.~\ref{tab:human-test-data}.

\begin{table}[t]
    \centering
\scalebox{0.8}{
    \begin{tabular}{llccc}
        \toprule
         & & \multicolumn{3}{c}{Captions} \\
         &  & \texttt{Synthesized} & \texttt{Human-Created} & Total \\
        \midrule
        \multirow{3}{*}{Map}
        & \texttt{Seen} & 15 & 15 & 30\\
        & \texttt{Unseen} & 15 & 15 & 30\\
        & Total & 30 & 30 & 60 \\
        \bottomrule
    \end{tabular}
    }
    \caption{Statistics on human-created trajectory data. Participants produced trajectory data with a carefully controlled set of synthesized/human-created captions and seen/unseen maps.}
    \label{tab:human-test-data}
\end{table}

\begin{table*}[t]
    \centering
    \scalebox{0.85}{
    \begin{tabular}{@{}lcccccccccccc@{}}
        \toprule
        & \multicolumn{6}{c}{Synthesized Captions} & \multicolumn{6}{c}{Human-created Captions} \\
        \cmidrule(lr){2-7} \cmidrule(lr){8-13}
            Models & R-1 & R-2 & R-L & BS-p & BS-r & BS-f1 & R-1 & R-2 & R-L & BS-p & BS-r & BS-f1 \\
            \midrule
            T5-Small & 0.080 & 0.020 & 0.066 & 0.529 & 0.520 & 0.539 & 0.055 & 0.006 & 0.047 & 0.602 & 0.593 & 0.613 \\
            T5-Base & 0.303 & 0.101 & 0.259 & 0.866 & 0.875 & 0.858 & 0.136 & 0.006 & 0.122 & 0.838 & 0.837 & 0.839 \\
            \midrule
            GPT-3.5 & 0.383 & 0.105 & 0.246 & 0.898 & 0.898 & 0.899 & 0.291 & \textbf{0.041} & 0.189 & 0.877 & 0.880 & 0.875 \\
            GPT-4 & 0.376 & 0.097 & 0.234 & 0.897 & 0.894 & 0.900 & \textbf{0.309} & 0.037 & 0.188 & 0.878 & 0.877 & \textbf{0.879} \\
            Llama2 & 0.389 & 0.137 & 0.272 & 0.886 & 0.876 & 0.898 & 0.254 & 0.032 & 0.163 & 0.861 & 0.855 & 0.868 \\
            \midrule
            Ours w/ 8 paraphrase& \textbf{0.436} & \textbf{0.163} & \textbf{0.329} & \textbf{0.914} & \textbf{0.920} & \textbf{0.907} & 0.306 & \textbf{0.041} & \textbf{0.205} & \textbf{0.883} & \textbf{0.890} & 0.876 \\
        \bottomrule
    \end{tabular}
    }
    \caption{Performance comparisons between synthesized and human-created captions on real human trajectories.}
    \label{tab:caption_result}
\end{table*}

\begin{table*}[t]
    \centering
    \scalebox{0.85}{
    \begin{tabular}{@{}lcccccccccccc@{}}   
        \toprule
        & \multicolumn{6}{c}{Seen Store Map} & \multicolumn{6}{c}{Unseen Store Map} \\
        \cmidrule(lr){2-7} \cmidrule(lr){8-13}
                Models & R-1 & R-2 & R-L & BS-p & BS-r & BS-f1 & R-1 & R-2 & R-L & BS-p & BS-r & BS-f1 \\
                \midrule
                T5-Small & 0.054 & 0.008 & 0.047 & 0.537 & 0.528 & 0.547 & 0.081 & 0.018 & 0.065 & 0.594 & 0.584 & 0.605 \\
                T5-Base & 0.220 & 0.055 & 0.192 & 0.851 & 0.855 & 0.848 & 0.219 & 0.052 & 0.189 & 0.852 & 0.856 & 0.849 \\
                \midrule
                GPT-3.5 & 0.344 & 0.079 & 0.224 & 0.888 & 0.890 & 0.887 & 0.329 & 0.067 & 0.210 & 0.887 & 0.887 & 0.887 \\
                GPT-4 & 0.346 & 0.070 & 0.215 & 0.889 & 0.887 & 0.890 & 0.339 & 0.064 & 0.207 & 0.886 & 0.884 & 0.888 \\
                Llama2 & 0.330 & 0.091 & 0.225 & 0.875 & 0.869 & 0.883 & 0.314 & 0.077 & 0.211 & 0.872 & 0.862 & 0.883 \\
                \midrule
                Ours w/ 8 paraphrase & \textbf{0.379} & \textbf{0.109} & \textbf{0.273} & \textbf{0.900} & \textbf{0.907} & \textbf{0.893} & \textbf{0.364} & \textbf{0.095} & \textbf{0.260} & \textbf{0.897} & \textbf{0.904} & \textbf{0.890} \\
        \bottomrule

    \end{tabular}
    }
    \caption{Performance comparisons between seen and unseen store maps on real human trajectories.}
    \label{tab:map_result}
\end{table*}

{
\tabcolsep = 1pt
\renewcommand{\arraystretch}{0.9}
\begin{table*}[h!t]
    \centering
    \scalebox{1}{
    \begin{tabular}{>{\centering\arraybackslash}p{4.6cm}p{0.9cm}p{10cm}}
        \toprule
        \multicolumn{3}{l}{\fontsize{9pt}{0pt}\selectfont{\textbf{Success Case 1 (Synthesized trajectories and captions)}}}\\
        \midrule
            \multirow{4}{*}[-0.1cm]{\adjustbox{valign=c}{\includegraphics[width=4.5cm]{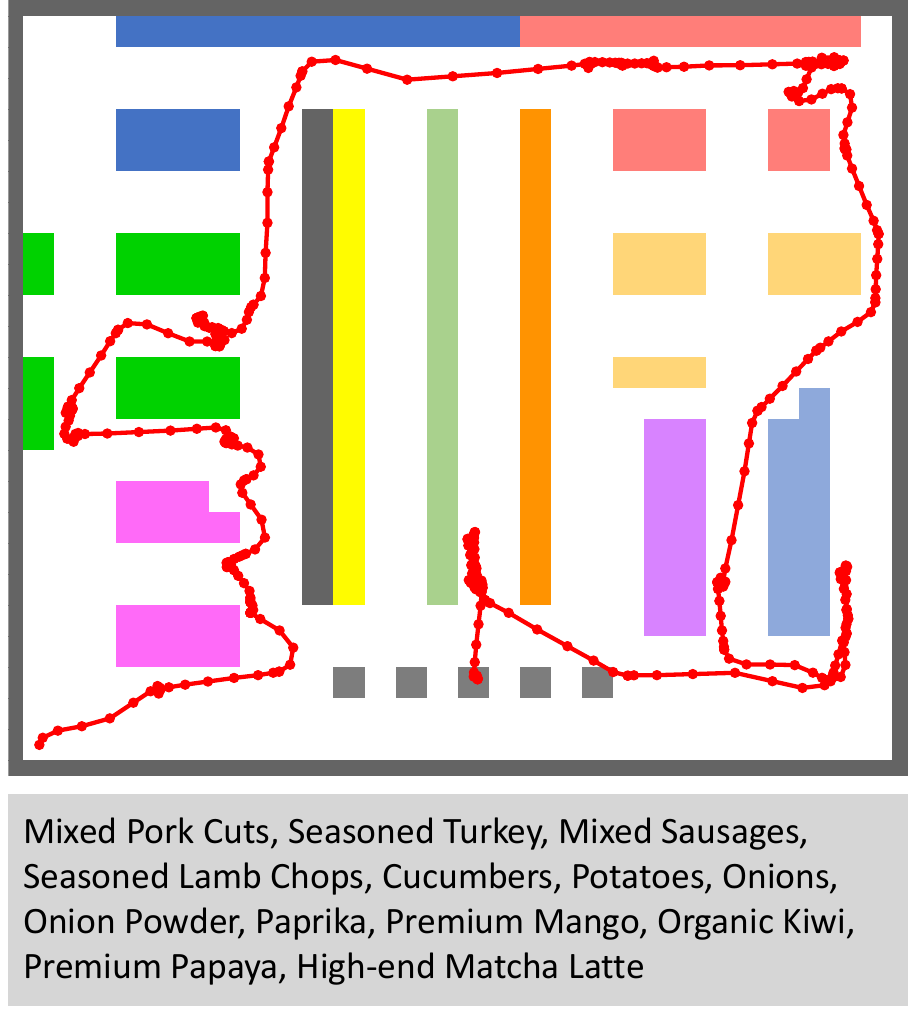}}} & \fontsize{9pt}{0pt}\selectfont{Label} &  \fontsize{9pt}{0pt}\selectfont{Customer is shopping for plant-based items. The customer has a predetermined list of vegan products like fruits, vegetables, grains, and dairy alternatives. The customer values quality over price and prefers organic and fresh items.} \\
        \cmidrule(lr){2-3}
        & \fontsize{9pt}{0pt}\selectfont{GPT-4} & \fontsize{9pt}{0pt}\selectfont{The customer is a health-conscious shopper who prefers organic and high-quality products. They usually shop for fresh fruits and vegetables, and they also enjoy premium snacks like mixed nuts. They prefer to season their food themselves, often purchasing basic seasonings like salt and pepper. They also enjoy high-end green tea, indicating a preference for natural and healthy beverages. This customer values quality over quantity and is willing to pay a premium for organic and luxury items.}\\
        \cmidrule(lr){2-3}
        & \fontsize{9pt}{0pt}\selectfont{Ours} & \fontsize{9pt}{0pt}\selectfont{A customer who is conscious of their health and enjoys eating fresh and unprocessed items. They have a predetermined list of purchases, which includes fresh fruits, vegetables, and grains. The customer prioritizes quality over price and prefers fresh and unprocessed items.}\\
        \bottomrule
        \toprule
        \multicolumn{3}{l}{\fontsize{9pt}{0pt}\selectfont{\textbf{Success Case 2 (Human-created trajectories and synthesized captions)}}}\\
        \midrule
            \multirow{4}{*}[-0.23cm]{\adjustbox{valign=c}{\includegraphics[width=4.5cm]{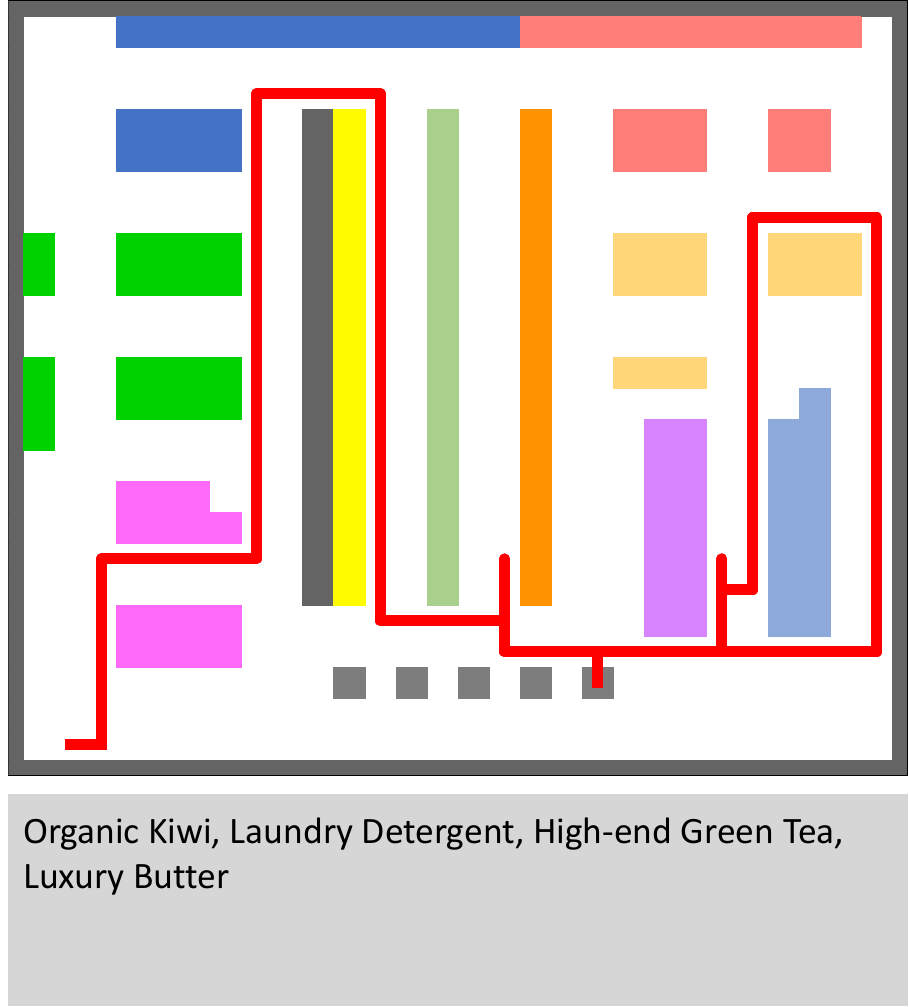}}} & \fontsize{9pt}{0pt}\selectfont{Label} &  \fontsize{9pt}{0pt}\selectfont{An environmentally conscious customer who prioritizes sustainable and eco-friendly products. The customer usually comes with a predetermined list of purchases, which includes items like organic produce, fair-trade coffee, and eco-friendly cleaning products. The customer is willing to pay more for products that align with their values and prefer unprocessed goods.} \\
        \cmidrule(lr){2-3}
        & \fontsize{9pt}{0pt}\selectfont{GPT-4} & \fontsize{9pt}{0pt}\selectfont{A health-conscious and high-end consumer who values quality over quantity. The customer usually purchases organic fruits, high-end beverages, and luxury dairy products. The customer also buys household goods, indicating a preference for maintaining a clean and organized home. The customer's shopping habits suggest a preference for luxury items and a willingness to pay more for high-quality products.}\\
        \cmidrule(lr){2-3}
        & \fontsize{9pt}{0pt}\selectfont{Ours} & \fontsize{9pt}{0pt}\selectfont{A customer who is environmentally conscious and values eco-friendly products. They have a predetermined list of purchases, focusing on eco-friendly products and eco-friendly alternatives. Quality is their primary concern, and they prefer eco-friendly products.}\\
        \bottomrule
        \toprule
        \multicolumn{3}{l}{\fontsize{9pt}{0pt}\selectfont{\textbf{Failure Case (Human-created trajectories and captions)}}}\\
        \midrule
            \multirow{4}{*}[-0.1cm]{\adjustbox{valign=c}{\includegraphics[width=4.5cm]{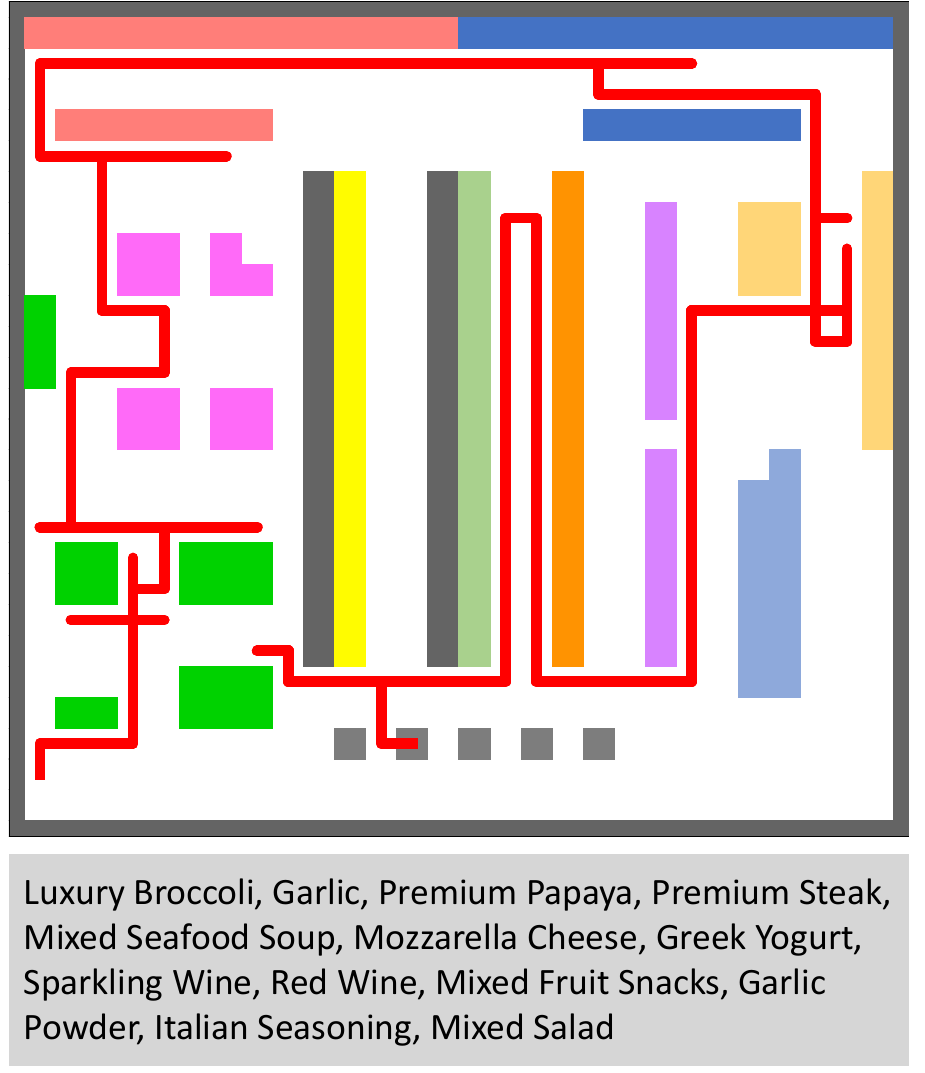}}} & \fontsize{9pt}{0pt}\selectfont{Label} &  \fontsize{9pt}{0pt}\selectfont{Newly married couple customer. They both want to enjoy cooking, so they look at and buy from a wide range of genres. They decide to a certain extent which days are fish-centered, meat-centered, and so on. If the main focus is fish, they buy more vegetables, and if the main focus is meat, they buy wine and other items together. Since they both work, they do not compromise on price.} \\
        \cmidrule(lr){2-3}
        & \fontsize{9pt}{0pt}\selectfont{GPT-4} & \fontsize{9pt}{0pt}\selectfont{A gourmet food enthusiast who enjoys cooking and experimenting with different recipes. They often shop for a variety of fresh vegetables, premium meats, and seafood. They also have a preference for luxury fruits and dairy products. They enjoy pairing their meals with a selection of wines and often indulge in snacks. They usually shop with a specific recipe in mind, often inspired by international cuisines.}\\
        \cmidrule(lr){2-3}
        & \fontsize{9pt}{0pt}\selectfont{Ours} & \fontsize{9pt}{0pt}\selectfont{A customer who is passionate about food and is willing to pay a premium for fresh produce, meats, and dairy products. Their shopping list consists of a mixture of fresh produce, meats, and dairy products. They are willing to pay a premium for fresh produce and are willing to pay a premium for quality.}\\
        \bottomrule
    \end{tabular}
    }
    \caption{Qualitative comparisons of ground-truth and generated captions. The movement trajectory is shown as a solid red line, with the purchase list displayed below. Colored rectangles represent shelves for different categories of items.}
    \label{tab:qualitative_result}
\end{table*}
}

\paragraph{[RQ.2] Can the model generalize to human-created trajectories/captions?}
Tab.~\ref{tab:caption_result} shows the quantitative results for human trajectories data, compared between when ground-truth captions are synthesized or created by human participants. 
Here we evaluated T5-Small and T5-Base, GPT-3.5/GPT-4/Llama2 each with 4 examples for in-context learning, and our captioning model with 8 paraphrases based on the previous result. Overall, our captioning model generalized well to those human-created data, with acceptably slight degradation of performances. Again, our model demonstrates comparable performance to GPT-3.5/4 and Llama2 despite its much smaller number of parameters. It is inevitably difficult to match generated captions with human-created ground truths exactly at word/phrase levels, as indicated by degraded ROUGE scores.  Nevertheless, the semantic consistency measured by BERT Scores remains as high as that for synthesized captions, indicating the practical usability.

\paragraph{[RQ.3] Can the model generalize to unseen maps?}
Tab.~\ref{tab:map_result} compares the performance between when store maps are seen (\ie, identical to those for training data) and unseen. For all models, we confirmed negligible performance degradation. This is practically beneficial, as major retailers often operate multiple stores that can have different layouts and item availability, where captioning systems should be easy to deploy.

\subsection{Qualitative Results and Failure Cases}
Tab.~\ref{tab:qualitative_result} illustrates some qualitative results of success and failure cases. If successful, our captioning model provides an accurate background context such as \emph{``have a list of items to buy''} (success cases 1 and 2). Based on additional information attached to items such as price and quantity, it is also possible to predict customer's preference, \eg, \emph{``prioritizes quality over price and prefers fresh and unprocessed items''} (success case 1) and \emph{``customer who is environmentally conscious''} (success case 2, against ground-truth label: \emph{``environmentally conscious customer''}).

Human-created captions can sometimes include demographic information of target individuals, such as \emph{``newly married couple customer,''} which are difficult to predict. It is also hard to generate sentences like \emph{``If the main focus is fish, they buy more vegetables, and if the main focus is meat, they buy wine and other items together.''} (the ground-truth label in the failure case). Still, our model appropriately infers the customer's preference, \eg, \emph{``willing to pay a premium for quality''} and \emph{``customer who is passionate about food''}. Additionally, unlike success cases 1 and 2, our model does not mention that the customer has the predetermined item list. This is consistent with the redundant trajectory of the failure case, suggesting that our model correctly inferred the customer's decision-making tendencies.

\subsection{Limitations and Practical Implications}
Our approach has a few limitations. As we obtain a captioning model by fine-tuning pretrained language models, its text generation capability would inevitably rely on that of the base model. Namely, our model cannot handle extremely long shopping activities beyond the maximum input token length for the base model. Moreover, there is no guarantee that the model won't hallucinate contexts that are totally irrelevant to a target individual. In practical system setup, it is crucial to post-process model outputs, for example, based on heuristic rules or manual inspection, so as not to present inappropriate captions to users. Recent work that seeks to mitigate hallucination~\citep{mundler2024self} would also help. Finally, similar to web search engines, it is necessary to allow for an opt-out option on the customer side for the use of inferred contextual captions in practical applications.

%% file: contents/6_related_works.tex
\section{Related Work}

\paragraph{Human movement analysis.}
Studies on human movements can be found in various research contexts, such as urban engineering~\citep{Pappalardo2016-vh, ASKARIZAD2020102687}, traffic simulation~\citep{Doniec2008-ai,Duives2013-pi}, autonomous driving~\citep{Camara2021-ga}, tourism~\citep{Li2018-fa, Payntar2021-dc}, and public health~\citep{Kraemer2020-tk}. Concrete techniques include pattern mining~\citep{Lam2017-hc,ghose2019mobile}, semantic mining~\citep{Parent2013-tx}, trajectory prediction~\citep{Rudenko2020-fp}, and crowd analysis~\citep{zhou2020understanding}. Compared to these prior arts, our work is the first to explore the potential of recent progress in large language modeling to empower human movement analysis and its application to retail scenarios.

\paragraph{Human activity captioning.}
Captioning human activities has been addressed mainly in computer vision, as a part of image captioning~\citep{hossain2019comprehensive} and video captioning~\citep{aafaq2019video}. Continuous efforts have been made to develop large-scale multimodal datasets that involve human activity data and their captions~\citep{krishna2017dense,grauman2023ego}. Nevertheless, much recent work seeks to exploit rich representations of human activities in visual data, which is not applicable to our task where only location trajectories and limited semantic information are available.

\paragraph{Generative models as data generators.}
Finally, there is a growing trend to utilize generative models to construct synthetic datasets. For example, generative adversarial networks and diffusion models have been used in computer vision to create or augment visual training data~\citep{Karras2019-fn,pmlr-v162-nichol22a}. LLMs have been used more widely for dataset generation, such as generating annotations~\citep{Feng2021-nl,zhang-etal-2023-llmaaa, Flamholz2024-ch, noauthor_undated-fu}, ranking~\citep{Hou2024-he,Qin2023-nl,Sun2023-jb}, and textual datasets~\citep{chen-etal-2023-places, Chung2023-cc}. Some recent work uses LLMs as virtual agents that produce realistic behaviors in simulated worlds~\citep{Park2023-ke,Kaiya2023-rv}. Our data synthesis framework is unique in terms of integrating LLMs and trajectory planners to produce diverse captioned human trajectories.

%% file: contents/7_conclusion.tex
\section{Conclusion}
\label{sec:conclusion}
We presented a new task named contextual captioning of human movement trajectories, and a dedicated learning-by-synthesis framework, \ie, \methodname, with a particular focus on retail scenarios. 
We leverage LLMs to synthesize realistic and diverse collection of contextual captions as well as concrete trajectories on store maps.  Our captioning model fine-tuned on these synthesized data demonstrated equal or even better performance than existing LLMs with a higher number of parameters. Moreover, the model well generalizes to human-created trajectories and captions.

Although this work focused exclusively on retail scenarios, we believe that the proposed task and framework would open up a new opportunity for adopting neural language generation techniques to various applications that need automated human activity understanding. This also raises new technical challenges such as effective encoding of very long trajectory data as input to language models and efficient inference of learned models to enable online captioning.

%% file: contents/appendix.tex
\clearpage
\onecolumn
\section{Prompt for Data Synthesis}
\label{sec:prompt}
\begin{figure*}[th]
    \centering
    \begin{tcolorbox}[title=STEP 1: Instruction for generating each contextual caption $S$]
    \small
        System:
        Your task is to generate descriptions of various customer intentions within a supermarket environment, elucidating their purchasing preferences and habits meticulously.\\

        Human:
        Kindly generate \{samples\} unique descriptions of customer intentions, ensuring each one is varied, embodying a range of customer profiles and shopping objectives. Every description should be comprehensively structured to include the following components:\\

        \begin{itemize}
            \item Outline the overarching characteristics defining the customer's shopping intention.
            \item Identify the categories of products the customer is likely to purchase or abstain from, such as a preference for meat over seafood, or vegetables over fruits.
            \item Clarify whether the customer arrives with a predetermined list of purchases or if they are likely to explore and decide while shopping.
            \item Elaborate on the customer's family structure,such as being a single individual, a couple, or part of a larger family, and how this influences their purchasing decisions.
            \item Highlight customer's preferences regarding the price and quality of products, specifying if they lean towards high-end items, discounted quality goods, or more affordable, lower-quality products.
            \item Describe the customer's preferences concerning the state of the products, such as pre-cut, seasoned, etc.
            \item If there is a dish the customer would like to cook, describe it. If not, please state that you do not.
            \item It is imperative to maintain strong consistency between the customer's "intention" and "num\_item\_to\_buy". For example, a family of five might buy a lot of items at once. These customers usually buy in bulk, getting many products in one visit. On the other hand, some customers come to the supermarket often, but they only buy a few things each time.
            \item Ensuring a close alignment between a customer's "intent" and their 'purchase\_consideration' is crucial. For instance, customers who are uncertain about their purchase choice or who explore various options typically exhibit a higher level of "purchase\_consideration". In contrast, customers who have a pre-determined purchase decision before visiting the store usually show lower "purchase\_consideration".
        \end{itemize}     
        Rule:\\
        Ensure all responses maintain the prescribed format and diversity in customer intentions is robustly represented!
        You must persist in generating sentences without cessation until you have produced at least \{samples\} intentions in total!!!\\

        Example:
    \end{tcolorbox}
    \caption{Prompt used for Step 1 in the Text2Traj phase.}
    \label{fig:prompt-step1}
\end{figure*}

\begin{figure*}[thpb]
    \centering
    \begin{tcolorbox}[title=STEP 2: Instruction for generating an abstract action plan consistent with each contextual caption generated in STEP 1.]
    \small
        System:
        As an adept AI, your task is to create a shopping plan for a customer, using their stated intentions, the total number of items they intend to purchase, and a provided list of product categories.\\

        Human:
        Your role is to allocate the total number of items the customer plans to purchase across the given product categories. This allocation should form a cohesive plan that aligns with the customer's intentions and preferences.\\

        Rule:
        Ensure all responses maintain the prescribed format!
        The total number of items in the shopping plan should be approximately \{num\_items\}.
        The distribution of products across categories must closely align with the customer's intention.\\

        \# Customer's intention
        \{intention\}\\

        \# category List
        \{category\_list\}

        \{format\_instructions\}
    \end{tcolorbox}
    \caption{Prompt used for Step 2 in the Text2Traj phase.}
    \label{fig:prompt-step2}
\end{figure*}

\begin{figure*}[thpb]
    \centering
    \begin{tcolorbox}[title=STEP 3: Instruction for generating item lists.]
    \small
        System:
        As a proficient AI assistant, your task is to curate two lists of products that align with the customer's intentions. You have access to detailed information, including the customer's intentions, product descriptions, the quantities they plan to purchase, and their level of purchase consideration.\\

        Human:
        Your goal is to create two lists based on the provided information:
        1. "inclined\_to\_purchase": Products that the customer is highly likely to purchase.
        2. "show\_interest": Products the customer might consider purchasing or show interest in, taking into account both the customer's intentions and their "purchase\_consideration" score.\\

        Guidelines:
        \begin{itemize}
            \item Purchases are planned only for products in the \{category\} category.
            \item Ensure that the total number of products in the "inclined\_to\_purchase" list for the \{category\} category is approximately \{num\_purchase\_items\}.
            \item Ensure that the total number of products in the "show\_interest" list for the \{category\} category is less than \{num\_purchase\_items\}.
            \item Align the "inclined\_to\_purchase" items in the \{category\} category with the customer's intentions.
            \item Generate the "show\_interest" list by carefully considering both the customer's intentions and their "purchase\_consideration" score, which ranges from 1 to 5. If the purchase\_consideration score is low, focus on a smaller "show\_interest" list. Conversely, if the score is high, the "show\_interest" list can be more extensive but should remain below \{num\_purchase\_items\} in total.
        \end{itemize}

        Tips:
        \begin{itemize}
            \item Pay close attention to the item descriptions and customer intentions provided.
        \end{itemize}

        \#\#\# Customers intention
        \{intention\}

        \#\#\# "purchase\_consideration" (1-5)
        \{purchase\_consideration\}

        \#\#\# Item description
        \{item\_description\}

        \{format\_instructions\}
    \end{tcolorbox}
    \caption{Prompt used for Step 3 in the Text2Traj phase.}
    \label{fig:prompt-step3}
\end{figure*}

\clearpage